\documentclass[conference]{IEEEtran}
\usepackage[usenames,dvipsnames]{color}
\usepackage{graphicx}
\usepackage{float}
\usepackage{url}
\usepackage{multirow}
\usepackage{algorithm}
\usepackage{algorithmic}
\usepackage[noadjust]{cite}
\usepackage{dsfont}
\usepackage{amsmath}
\usepackage{authblk}

\IEEEoverridecommandlockouts
\newcommand{\SkyBlue}[1]{\textcolor{Black}{#1}}
\newcommand{\bx}[1]{\SkyBlue{\textsc{} #1}}

\newcommand{\PineGreen}[1]{\textcolor{Black}{#1}}
\newcommand{\pl}[1]{\PineGreen{\textsc{} #1}}

\newcommand{\Blue}[1]{\textcolor{Black}{#1}}
\newcommand{\hx}[1]{\Blue{\textsc{} #1}}

\newcommand{\Orange}[1]{\textcolor{Black}{#1}}
\newcommand{\lxd}[1]{\Orange{\textsc{} #1}}

\newcommand{\BrickRed}[1]{\textcolor{Black}{#1}}
\newcommand{\Yuyu}[1]{\BrickRed{\textsc{} #1}}

\begin{document}

\title{Combination of Diverse Ranking Models for Personalized Expedia Hotel Searches}

\author[,1]{Xudong Liu$^*$\thanks{$^*$ These authors equally contributed to this work.}}
\author[,2]{Bing Xu$^*$}
\author[,3]{Yuyu Zhang$^*$}
\author[,4]{Qiang Yan$^*$}
\author[,3]{Liang Pang$^*$}
\author[3]{Qiang Li}
\author[3]{Hanxiao Sun}
\author[,3]{Bin Wang$^\dag$\thanks{$^\dag$ Team advisor}}
\affil[1]{Institute of Automation, Chinese Academy of Sciences}
\affil[2]{University of Alberta}
\affil[3]{Institute of Computing Technology, Chinese Academy of Sciences}
\affil[4]{Taobao Inc.}
\affil[1,2,3]{\{lxdwinwin, antinucleon, zhangyuyu2008, pl8787\}@gmail.com}
\affil[4]{yanqiang.yq@taobao.com}
\affil[ ]{}
\affil[ ]{}

\maketitle

\begin{abstract}
\bx{The ICDM Challenge 2013 is to apply machine learning to the problem of hotel ranking, aiming to maximize purchases according to given hotel characteristics, location attractiveness of hotels, user’s aggregated purchase history and competitive online travel agency (OTA) information for each potential hotel choice. This paper describes the solution of team "binghsu \& MLRush \& BrickMover". We conduct simple feature engineering work and train different models by each individual team member. Afterwards, we use listwise ensemble method to combine each model's output. Besides describing effective model and features, we will discuss about the lessons we learned while using deep learning in this competition.}
\end{abstract}

\section{Introduction}
\bx{ICDM Challenge 2013 requires learning to rank hotels to maximize purchases for given hotel queries by Expedia.com. The dataset which is provided by Expedia.com, contains hotel characteristics, location attractiveness of hotels, user’s aggregate purchase history and competitive OTA information for each search\_id-hotel pair. Hotels for each user query are assigned relevance grades as following: 5 for user purchased a room; 1 for user clicked the information of the hotel and 0 for user neither click nor book. The data is split by organizers by randomly split.}

\bx{The training data contains 399,344 unique search lists and 9,917,530 points. The test data contains 266,230 search lists and 6,622,629 points. 25\% of the test data is used for evaluating in public leaderboard and the remaining 75\% is used as final private test data.\footnote{Complete leaderboard can be found at http://www.kaggle.com/c/expedia-personalized-sort/leaderboard}}

\bx{The evaluation metric for this competition is Normalized Discounted Cumulative Gain (NDCG), which is commonly used in ranking\cite{liu2009learning}. According to the announced result, our approach achieved $5^{th}$ on the private leaderboard with 0.53102 NDCG@38 score.}

\bx{The paper is organized as follows. Section 2 outlines the framework of our approaches. Section 3 discusses the preprocessing and feature engineering. Section 4 introduces the single effective models we use. Section 5 describes the ensemble experiments we use to boost our performance. Finally we conclude this paper and further discuss lessons we learned in Section 6.}

\section{Framework}
\bx{This section introduces the architecture and softwares we used in our system. Then it discusses the self-split internal validation set from the training set, which is important in model evaluating and combining different models. }

\subsection{System Overview}

\begin{figure}
\centering
\includegraphics[width=.44\textwidth]{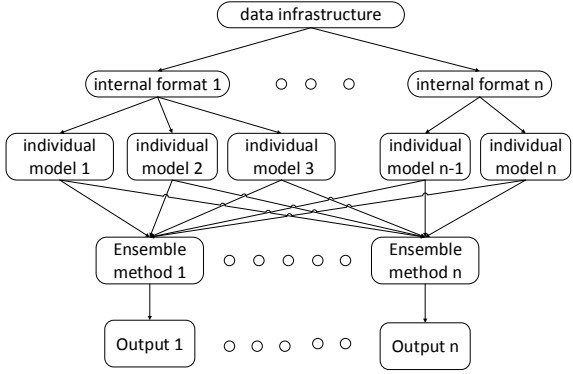}
\caption{The overall architecture of our approach.}
\label{architecture}
\end{figure}

\bx{Our system can be divided into three parts: data infrastructure, training individual models and ensemble as shown in Figure \ref{architecture}. The data infrastructure is based on pandas\cite{pandas}. In this step we use pandas to store data. And we do some feature engineering in this stage. The output of the data infrastructure can be in different format like numpy binary array, SVMRank\cite{svmrank} text format and LIBSVM\cite{libsvm} text format. In the second stage, we explore diverse approaches to generate various models, including logistic regression,
support vector machine, random forest, gradient boosting machine, factorization machine, LambdaMART and deep neural network. In the last step, we combine all different results on the internal validation set and test set by using listwise approach, linear approach and deep neural network approach.}

\bx{We use LIBLINEAR\cite{liblinear} and SVMRank\cite{svmrank}for pairwise logistic regression; Random Forest\cite{randomforest} from scikit-learn\cite{pedregosa2011scikit}; Ranking algorithms like AdaRank\cite{adarank}, LambdaMART\cite{lambdamart} from RankLib\footnote{http://sourceforge.net/p/lemur/wiki/RankLib/}. And we also use Gradient Boosting Machine\footnote{http://cran.r-project.org/web/packages/gbm/index.html}\cite{gbm}, Extremely Randomized Trees\footnote{http://cran.r-project.org/web/packages/extraTrees/index.html}\cite{ext} from R, deep neural network implementation from PyLearn2\cite{goodfellow2013pylearn2} and Factorization Machine libFM\footnote{http://www.libfm.org/}\cite{libfm}}.

\subsection{Internal Validation Set}
\bx{A validation set can be used to evaluate single model without submitting the test result to the leaderboard. And validation set is very important for combining different models. Usually, the training set can be divided in the ratio 6:2:2. But in this dataset the validation result is quite robust so we just keep 10\% data as the validation. We use the rule $srch\_id \% 10 == 1$ to generate validation set and others to be sub training set.}

\section{Preprocessing}
\subsection{Listwise Feature for Point/Pairwise Model}
\bx{Some of the feature's rank in the list is used as a single feature for each hotel choice. Here are the most important listwise ranking features. 
\begin{itemize}
\item price\_usd
\item prop\_starrating
\item prop\_location\_score2
\end{itemize}
Other listwise ranking features we proposed but had no time to evaluate including: rank of exp(prop\_log\_historical\_price)-price\_usd, rank of click/booking bias and so on. }

\bx{Listwise features is a bridge to bring listwise information to point/pair wise models.}

\subsection{Composite Features}
\bx{Composite features is a method that combine two different features. For example, now we combine $srch\_room\_count$ and $srch\_booking\_window$, the $count\_window$ feature equals $srch\_room\_count * \max(srch\_booking\_window) + srch\_booking\_window$. }

\subsection{Dealing Missing Feature Values}
\bx{There are many missing feature values in the data such as prop\_location\_score2. We use the first quartile calculated by the country which the data point located in to represent the missing data.}

\subsection{Use 10\% data}
\bx{We randomly sample 10\% of the data by srch\_id to generate new training data. Using the new training data we can train a model with very small difference from the model trained by total training data.}

\subsection{Use Balanced Data}
\bx{Balanced data is used in training random forest and deep neural network. For there are only 4.4\% positive data points among the 9.9 million data points, we choose one positive example and randomly choose one negative example. In this way we can train tree-based models with a large amount of trees in a reasonable time.}

\subsection{Split Data by prop\_country\_id}
\bx{Based on prop\_country\_id we split the data into 172 pieces and train independent models on each piece. This method greatly reduces time on training tree based models.}

\subsection{Use Bucket to Binarize Float Feature}
\bx{Bucket is a strong rounding method to binarize the float feature. The bucket algorithm can be described as Algorithm \ref{alg:bucket}. By using bucket, the float features are in smaller variance.}

\begin{algorithm}[h!]
\caption{$Bucket(feature, bucket\_number)$}\label{alg:bucket}
\begin{algorithmic}[Bucket]
\REQUIRE An integer $BUCKET > 0$\\
$description$ = \{\}\\
$binary\_feature$ = zeros(($feature.size, bucket\_number$))
\FOR{$i=1$ to $bucket\_number$}
\STATE $description[i] = feature.quantile(i / bucket\_number)$
\ENDFOR

\FOR{$i=0$ to $feature.size$}
\STATE $j=1$\\
\WHILE{$feature.at(i) < description[j]$}
\STATE $j = j + 1$
\ENDWHILE
\STATE $binary\_feature[i][j]=1$
\ENDFOR
\RETURN $binary\_feature$
\end{algorithmic}
\end{algorithm}

\section{Models}

\subsection{Logistic Regression}
\Yuyu{As a classical model for binary classification, logistic regression is used as our initial attempt in this competition. We tried both the binary logistic regression and the multinomial logistic regression, while the former one performs obviously better. So in this part, we will only introduce our approach on binary logistic regression.} 

\Yuyu{We firstly pre-process the data by merging the clicked and booked items within each query as positive instances, while all the left items are regarded as negative instances. With some feature engineering work, which will be discussed in details later, all the instances can be represented as a series of feature vectors. Now it becomes a standard binary classification problem, though the data here is very unbalanced since the negative instances are overwhelming. Therefore, we then adjust the weight in the cross-entropy error function (also known as negative log-likelihood function) to tackle the issue of data unbalance. The revised error function is shown as follows:}
\begin{equation}
\begin{split}
CEE &= -\sum_{n=1}^{N}\log[\mu_i^{\alpha\mathds{I}_{(y_i=1)}} \cdot (1-\mu_i)^{\mathds{I}_{(y_i=0)}}]  \\
&= -\sum_{n=1}^{N}[\alpha y_i\log\mu_i + (1 - y_i)\log(1 - \mu_i)]
\end{split}
\end{equation}
where $\alpha$ is the parameter of class weight decided by input data, $\mu$ is the output of sigmoid function representing the probability to be positive instance, $\mu_i$ is for the $i^{th}$ instance.

\Yuyu{Since the cross-entropy error above is convex, it has a unique global minimum, we use gradient descent approach to find the optimal weight vectors or model parameters. Surprisingly, although the model of logistic regression is simple and the model itself is designed for classification rather than object ranking, the performance is fairly good, which can achieve over 0.52 in terms of NDCG on public leaderboard.}

\subsection{Pairwise Logistic Regression}
\pl{In order to use full train set as pairwise in logistic regression model, we use FTRL-Proximal algorithm and liblinear\cite{liblinear} and SVMRank\cite{svmrank} build in function. The FTRL-Proximal algorithm \cite{ftrl}, can be seen as a hybrid of FOBOS and RDA algorithms, and significantly outperforms both on a large, realworld dataset.}

\lxd{We use the whole training set to train this model with simply seven features. The featurelist is srch\_id, prop\_id, srch\_destination\_id, prop\_starrating, prop\_location\_score1, prop\_location\_score2, price\_usd. And this single FTRL model archives 0.51273 NDCG@38 on validation set.}

\subsection{Random Forest}
\bx{After forming a split dataset in 172 pieces by using prop\_country\_id, we balanced each piece. For each unique balanced prop\_country\_id data piece we train an independent random forest model \cite{randomforest} with 3200 trees. With listwise ranking features, the 172 random forest models achieve nearly 0.51 NDCG@38 score in internal validation set. Some failed cases happened in predicting for some countries, so to make the example count of test and internal validation set equal we simply combine the score with a pairwise logistic regression trained by liblinear\cite{liblinear} model with 0.47 NDCG@38 score on validation, then the mixture balanced random forest model achieves nearly 0.52 NDCG@38 on validation set.}

\subsection{Gradient Boosting Machine}
\pl{Gradient Boosting Machine(GBM) \cite{gbm} is a machine learning technique for regression problems, which produces a prediction model in the form of an ensemble of weak prediction models, typically decision trees. It builds the model in a stage-wise fashion like other boosting methods do, and it generalizes them by allowing optimization of an arbitrary differentiable loss function. The gradient boosting method can also be used for classification problems by reducing them to regression with a suitable loss function.}

\pl{We use GBM in R language with the balanced data. The most 20 relevant feature show in Table~\ref{table:t1}. Fm\_score and Lr\_Score is the rank score that simple Factorization Machine and Linear Regression predicted using only visitor and query features. Date time is transform to unixstamp as a continuous feature. Feature name with the suffix \_cnt means one-way count of the feature in the combine of train and test set. The feature ump, price\_diff, starrating\_diff, per\_fee, score2ma,  total\_fee and score1d2 is generated by the formula below.
}
\small{$$ump = exp(prop\_log\_historical\_price)-price\_usd$$
$$price\_diff = visitor\_hist\_adr\_usd - price\_usd$$
$$starrating\_diff = visitor\_hist\_starrating - prop\_starrating$$
$$per\_fee = \frac{price\_usd*srch\_room\_count}{srch\_adults\_count+srch\_children\_count}$$
$$score2ma = prop\_location\_score2*srch\_query\_affinity\_score$$
$$total\_fee = price\_usd*srch\_room\_count$$
$$score1d2 = \frac{prop\_location\_score2+0.0001}{prop\_location\_score1+0.0001}$$}
\pl{This single GBM model archives 0.52477 NDCG@38 on validation set.}

\pl{Without one-way count features the GBM model archives 0.50099 NDCG@38 on validation set which is useful in ensemble process.}

\begin{table}
\centering
\caption{The top 20 relevance features.}\label{table:t1}
  \begin{tabular}{|c|c||c|c|}
    \hline
    \bfseries Feature & \bfseries Rel. & \bfseries Feature & \bfseries Rel. \\
    \hline\hline
    fm\_score & 50.35 & random\_bool & 1.00\\
    \hline
    lr\_score & 12.99 & srch\_destination\_id\_1\_cnt & 0.84\\
    \hline
    prop\_location\_score2 & 10.19 & date\_time & 0.63\\
    \hline
    ump & 3.58 & per\_fee & 0.61\\
    \hline
    price\_diff & 2.40 & price\_usd & 0.54\\
    \hline
    random\_bool\_1\_cnt & 2.26 & score2ma & 0.50\\
    \hline
    random\_bool\_f & 1.57 & prop\_location\_score2\_1\_cnt & 0.49\\
    \hline
    starrating\_diff & 1.39 & prop\_review\_score & 0.49\\
    \hline
    prop\_id\_1\_cnt & 1.25 & total\_fee & 0.47 \\
    \hline
    score1d2 & 1.06 & orig\_destination\_distance & 0.47 \\
    \hline
  \end{tabular}
\end{table}

\subsection{Extreme Randomized Trees}
\pl{Extremely Randomized Trees (ERT) is proposed by \cite{ext}.This method is similar to the Random Forests algorithm in the sense that it is based on selecting at each node a random subset of K features to decide on the split. Unlike in the Random Forests method, each tree is built from the complete learning sample (no bootstrap copying) and, most importantly, for each of the features (randomly selected at each interior node) a discretization threshold (cut-point) is selected at random to define a split, instead of choosing the best cut-point based on the local sample (as in Tree Bagging or in the Random Forests method).}

\pl{ERT model using in learning to rank task is proposed by \cite{l2rert}. We also use this model with the same feature set as GBM model and archives 0.51699 NDCG@38 on validation set.}

\subsection{Factorization Machine}
\lxd{Factorization Machine \cite{libfm} is widely used in Recommender System. It is also a kind of Regression model. So it can be used as a pointwise model. We did a lot of work in feature engineering and the single model archives 0.5171 NDCG@38 on validation set. As the feature engineering and the model itself bring much diverse, it works well when ensemble. Some of the features are listed in Tab.~\ref{table:t2}.}

\begin{table}
  \centering
  \caption{Features for Factorization Machine.}\label{table:t2}
  \begin{tabular}{|c|c|c|c|}
    \hline
    \bfseries Bin\_ID Feature & \bfseries Normalized Feature & \bfseries Ranking Feature \\
    \hline\hline
    prop\_id & price\_usd & price\_rank\\
    \hline
    srch\_destination\_id & prop\_location\_score1 & price\_diff\_rank\\
    \hline
    srch\_room\_count & prop\_location\_score2 & star\_rank\\
    \hline
  \end{tabular}
\end{table}

\lxd{For the model of Factorization Machine, our features are built in different ways. Some of the features are normalized (e.g. pro\_location\_score1), others are divided into bins (e.g. srch\_booking\_window). The ranking feature (e.g. price\_rank) means the rank value of the identical query, which works fairly well in this model. The key point of this model is that features should be used in the right way and ranking feature brings listwise information to this pointwise model.}
\subsection{LambdaMART}
\begin{figure}
\centering
\includegraphics[width=.44\textwidth]{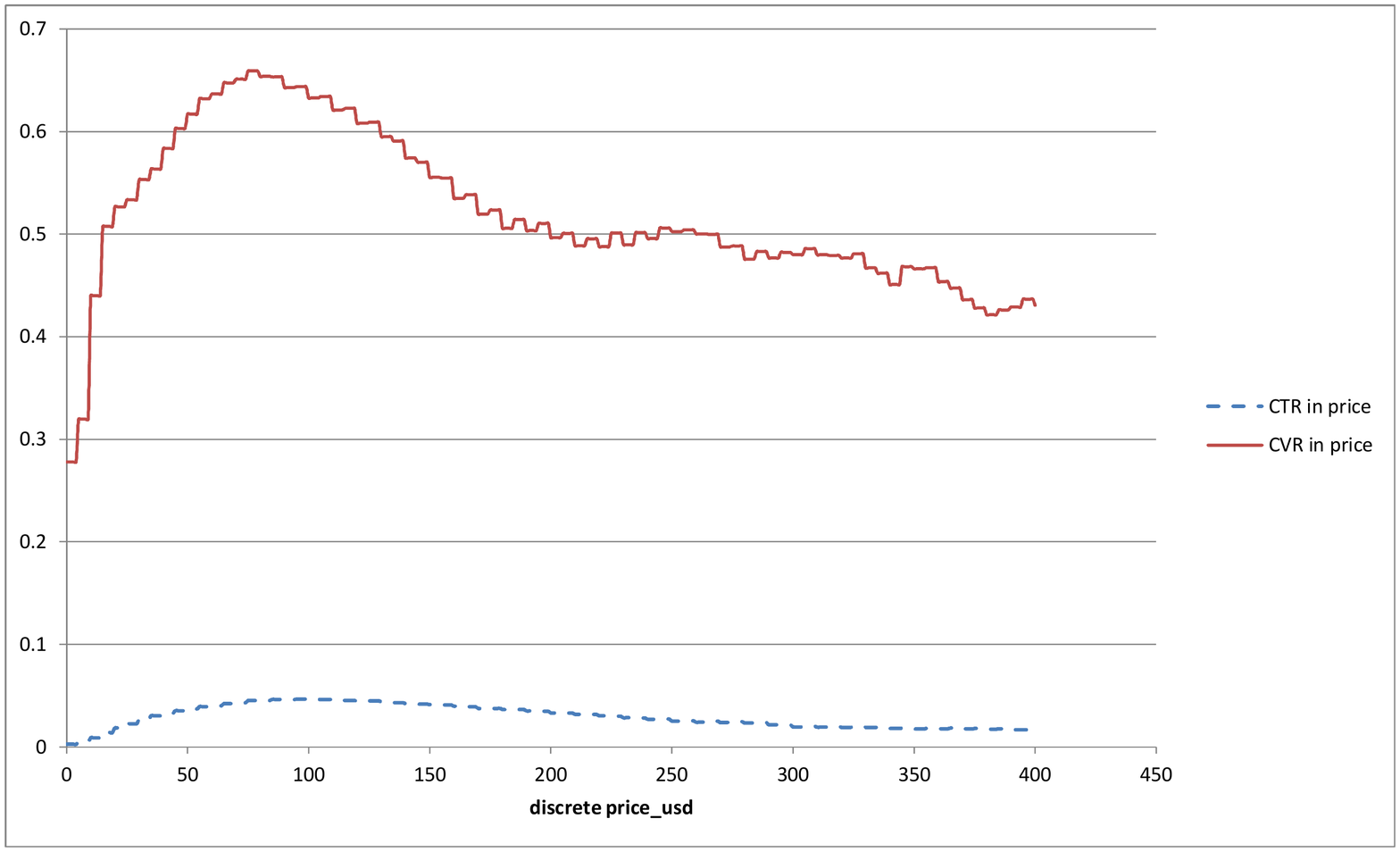}
\caption{The CTR and CVR in discrete price\_usd scale.}
\label{pricefea}
\end{figure}

\pl{LambdaMART model \cite{lambdamart} \cite{lambdamart1} with CTR(clickthrough rates Eq~\ref{eq:ctr}) and CVR(Booking rates Eq~\ref{eq:cvr}) features and original features in full train set. CTR and CVR are calculate in two scale, prop\_id and discrete price\_usd (Fig.~\ref{pricefea}). This model archives 0.51149 NDCG@38 on validation set.}

\begin{equation}\label{eq:ctr}
CTR_i = \frac{\#(Click_i)}{\#(Presentation_i)}
\end{equation}

\begin{equation}\label{eq:cvr}
CVR_i = \frac{\#(Booking_i)}{\#(Click_i)}
\end{equation}

\lxd{Two of the team members use LambdaMART independently. The other LambdaMART model is based on normalized features, ranking features and result features. Some features are listed in Tab.~\ref{table:t3} and this single LambdaMART model archives 0.5243 NDCG@38 on validation set}
\begin{table}
  \centering
  \caption{Features for LambdaMART.}\label{table:t3}
  \begin{tabular}{|c|c|c|c|}
    \hline
    \bfseries Normalized Feature & \bfseries Ranking Feature & \bfseries Result Feature\\
    \hline\hline
    prop\_starrating & price\_rank & fm\_score\\
    \hline
    prop\_location\_score1 & price\_diff\_rank & lr\_score\\
    \hline
    prop\_location\_score2 & star\_rank & \\
    \hline
  \end{tabular}
\end{table}
\lxd{All these features except fm\_score and lr\_score are introduced in the Factorization Machine part. Fm\_score and lr\_score are learnt by visitor and query features. These features won't work in pairwise or listwise models as they have the same value in one query. But in this way, these features contributes its bias in pairwise or listwise models. }

\subsection{Deep Learning Approach}
\begin{figure}
\centering
\includegraphics[width=.44\textwidth]{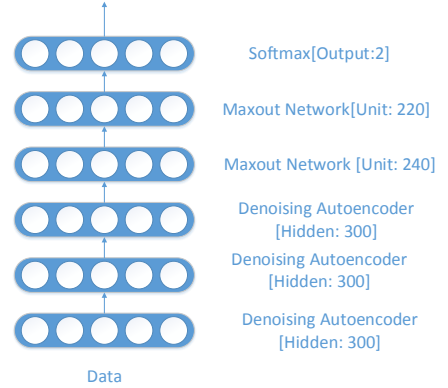}
\centering
\caption{The architecture of deep neural network.}
\label{dnn}
\end{figure}

\bx{Deep Learning has been successfully used in many fields such as image processing and acoustic processing. It is an open question to know whether the complex network and non-convex model can improve ranking and recommendation system. In practice, we first transform selected features into binary by using Algorithm \ref{alg:bucket}. To make it simpler, we choose to train independent models for each country by using same network framework in the figure. \ref{dnn}. We choose Maxout network\cite{maxout} because it provides much more regularization than other network layer. We discover the following situations while pretraining the denoising autoencoders:
\begin{enumerate}
\item Reconstruction error fixed around a large number (in most of cases)
\item Reconstruction error gradually become small (especially for prop\_country\_id=219)
\end{enumerate}
As the result shown before, composite feature is important. We want to use 3 layer denoising autoencoder to find at least 3-level and robust composite features. But for most of cases, the training data is not enough to make training process jump out of local minimum. And for the largest No.219 model, although pretraining looks fine, but while the training, the error rate on validation sucks to 4.3\%, which is quite near the positive ratio of the data. The mean of parameters of the softmax layer stay in less than $10^{-4}$. It means this deep neural network can not accurate predict the unbalanced data. But it achieves 0.48 NDCG@38 on local validation set, which means it learns something out. After switching to balanced data, the training process is a little smooth. The error rate reduces from 50\% to near 10\%. But the local NDCG@38 score only improve to near 0.49. Vertical ensemble\cite{hesm} helps improve a little but still can not make deep network as a strong single model. Linear embedding with other models can improve 0.004 NDCG@38 score on the local validation set of No. 219. We will provide further discussion in later chapter.}

\section{Ensemble Models}
\subsection{z-score}
\hx{As mentioned, we split the test set into two parts, by query. $10 \%$ of the data served for a validation set, which helps us to choose and combine rankers. We experimented with several linear methods to combine and the linear combining is nothing more than selecting several effective rankers and assign a set of parameters to them that the combination can achieve a better result than ever single model.}

\hx{The simplest, also work best, method we try was z-score ensemble Eq~\ref{Eq.1}. We added corresponding score in each ranker into the ensemble ranker in the beginning, but find it was not working well. For the rankers were derived from different models, the scope of their scores vary greatly. It is unreasonable to simply add them without normalization, so we employed z-score to normalize the rank scores.}

\begin{equation}\label{Eq.1}
Z(x)=\frac{x-\bar{x}}{\sigma(x)}=\frac{x-\bar{x}}{\sqrt{(x-\bar{x})^2/n}}
\end{equation}

\hx{We tried two ways of z-score, global z-score and query z-score. For the reason that we just compared the scores in each query, we calculated the z-score of every each query.}

\hx{In addition, instead of using greedy search, we tuned parameter manually. Lacking of a local test set to do cross validation and for fear of overfitting on local test set, we have to do it manually. Manually tuning could avoid excessive parameters, so, to some extent, it is a way of simple normalization, even not beautiful enough. Global z-score was the way we finally choose to ensemble and also the way achieve the highest NDCG score on private board.}

\subsection{GBM Ensemble}
\pl{Treat each model's output on local test set as a feature of GBM model \cite{gbm}. The target to learn is the click ground truth in local test set. In order to get a better result, we also include some significant features, such as prop\_location\_score1, prop\_location\_score2 and price\_usd.}

\pl{We use 30 models and 120 trees, and it archives 0.53573 NDCG@38 on the local test set, but only 0.53053 NDCG@38 on online test set. }
\subsection{Deep Learning}
\bx{We want to check whether the deep composite of the models output can make any progress. By using models with local NDCG score of 0.505, 0.508, 0.510, 0.511, 0.512, 0.513,0.519, 0.521. With ReLU\cite{relu} network or Maxout network, our model achieves around 0.526 on private leaderboard. And dropout logistic regression achieves best result of 0.52729 NDCG@38 on private leaderboard. They are still weaker than other ensemble methods.}

\subsection{Listwise Ensemble}
\bx{
Previous ensemble methods don't involve the listwise information. So we are trying to use LambdaMART to ensemble all the models. By using z-score normalization, it achieves our final score: 0.53249 NDCG@38 on public leaderboard, and 0.53102 on the final private leaderboard.}

\section{Conclusion and Further Discussion}
\subsection{Conclusion}
\Yuyu{In this paper, we present our approaches in ICDM Contest 2013. Due to the large size of data provided by Expedia, we use both random sampling and balanced sampling methods to construct reliable validation set. Diverse ranking models are described in this paper, including modified logistic regression, random forest, gradient boosting machine, extreme randomized trees, factorization machine, and lambdaMART. We also attempt to adopt deep learning approach, which will be further discussed in next sub-section. With these individual ranking models, we introduce our ensemble methods to combine individual models, including z-score, GBM, deep learning and listwise ensemble. The combination of models significantly improves the ranking accuracy in terms of NDCG on both public and private leaderboards.}

\subsection{Lessons Learned in Deep Learning}
\bx{Based on the split data, there is a serious problem of lacking training data. And reconstruction error may not reflect the best optimization direction. We suggest in later practice it is better to check whether pretraining learns out distributed representations other than only rely on the reconstruction error. Bucket may not be a optimized solution to normalize feature input A better normalization may lead to better generalization capacity for the deep networks.}

\section*{Acknowledgment}
We thank to Prof. Russ Greiner from University of Alberta for his advice while writing this report. And we also thank to Prof. Hong Hu from Institute of Computing Technology for his advice on data analysis.

\end{document}